\pdfoutput=1

\documentclass[11pt]{article}

\usepackage[]{acl}

\usepackage{times}
\usepackage{latexsym}

\usepackage[T1]{fontenc}

\usepackage[utf8]{inputenc}

\usepackage{microtype}

\usepackage{algorithm}
\usepackage{algorithmic}
\usepackage{siunitx}    
\usepackage{relsize}
\usepackage{etoolbox}
\robustify\smaller

\usepackage{newfloat}
\usepackage{listings}
\floatstyle{ruled}
\newfloat{listing}{tb}{lst}{}
\floatname{listing}{Listing}

\usepackage{soul}
\usepackage{amsfonts}
\usepackage{multirow}

\usepackage{dsfont}
\usepackage{commath}

\usepackage{amsmath}
\usepackage{mathtools}

\usepackage{bbold}

\usepackage[utf8]{inputenc}
\usepackage{pgfplots}
\DeclareUnicodeCharacter{2212}{−}
\usepgfplotslibrary{groupplots,dateplot}
\usetikzlibrary{patterns,shapes.arrows}
\pgfplotsset{compat=newest}

\usepackage{placeins}
\usepackage{subcaption}
\usepackage{mwe}
\usepackage{pifont}
\usepackage{xcolor}
\usepackage{paralist}

\usepackage{booktabs}
\usepackage{relsize}
\usepackage{array}

\newcolumntype{V}{>{\smaller}l}

\newcommand{\ex}[1]{{\small\texttt{#1}\xspace}}

\newcommand{\SysName}{\textsl{fairlib}\xspace}

\newcommand{\class}[1]{\textsf{#1}\xspace}
\newcommand{\AAE}{\class{AAE}}
\newcommand{\SAE}{\class{SAE}}

\newcommand{\dataset}[1]{\textsc{#1}\xspace}
\newcommand{\Moji}{\dataset{Moji}}
\newcommand{\Bios}{\dataset{Bios}}

\newcommand{\method}[1]{\textsc{#1}\xspace}
\newcommand{\Standard}{\method{Standard}}
\newcommand{\INLP}{\method{INLP}}
\newcommand{\Adv}{\method{Adv}}
\newcommand{\EAdv}{\method{EAdv}}
\newcommand{\DAdv}{\method{DAdv}}
\newcommand{\AAdv}{\method{AAdv}}
\newcommand{\ADAdv}{\method{ADAdv}}
\newcommand{\Gate}{\method{Gate}}

\newcommand{\BD}{\method{BD}}
\newcommand{\CB}{\method{CB}}
\newcommand{\JB}{\method{JB}}
\newcommand{\EO}{\method{BTEO}}
\newcommand{\GDDiff}{\method{EO$_\mathrm{{CLA}}$}}

\newcommand{\FairBatch}{\method{FairBatch}}
\newcommand{\FairSCL}{\method{FairSCL}}

\newcommand{\tp}{\ensuremath{\text{TP}}}
\newcommand{\tn}{\ensuremath{\text{TN}}}
\newcommand{\fp}{\ensuremath{\text{FP}}}
\newcommand{\fn}{\ensuremath{\text{FN}}}

\newcommand{\DTO}{\ensuremath{\text{DTO}}\xspace}

\usepackage{bm}

\usepackage{xspace}
\usepackage{adjustbox}

\definecolor{applegreen}{rgb}{0.55, 0.71, 0.0}

\definecolor{myBlue}{rgb}{0.12156863, 0.46666667, 0.70588235}
\definecolor{myOrange}{rgb}{1., 0.49803922, 0.05490196}
\definecolor{myRed}{rgb}{1,0,0}

\usepackage{mathtools, caption}

\usepackage{circuitikz}
\usetikzlibrary{external}
\usepackage[frozencache,cachedir=.]{minted}

\title{\SysName: A Unified Framework for Assessing and Improving Classification Fairness
}

\author{
Xudong Han$^{1}$ \qquad Aili Shen$^{1}$ \qquad Yitong Li$^{2}$ \qquad Lea Frermann$^{1}$ \\  {\bf Timothy Baldwin$^{1,3}$}  {\bf Trevor Cohn$^{1}$}\\ 
$^{1}$The University of Melbourne \\ 
$^2$Huawei Technologies Co., Ltd. \\
$^{3}$MBZUAI \\ 
\url{xudongh1@student.unimelb.edu.au}, \url{{aili.shen,lfrermann,tbaldwin,t.cohn}@unimelb.edu.au}
}

\begin{document}
\maketitle
\begin{abstract}
This paper presents \SysName, an open-source framework for assessing and improving classification fairness. It provides a systematic framework for quickly reproducing existing baseline models, developing new methods, evaluating models with different metrics, and visualizing their results.
Its modularity and extensibility enable the framework to be used for diverse types of inputs, including natural language, images, and audio.
In detail, we implement 14 debiasing methods, including pre-processing,
at-training-time, and post-processing approaches. 
The built-in metrics cover the most commonly used fairness criterion and can be further generalized and customized for fairness evaluation.
\end{abstract}

\section{Introduction}

While neural methods have achieved great success for classification, it has been shown that naively-trained models often learn spurious correlations with protected attributes like user demographics or socio-economic factors, leading to allocation harms, stereotyping, and other representation harms to users~\citep{badjatiya2019stereotypical, zhao2018gender, li-etal-2018-towards, diaz2018addressing, wang2019balanced}.

Various bias evaluation metrics have been introduced in previous studies to gauge different biases.
One common way of bias evaluation calculates the differences (``GAP'') between subgroup performances, such as equal opportunity and predictive equality, which evaluate the gap in the True Positive Rate (TPR) and True Negative Rate (TNR) respectively, across different protected attributes~\citep{de2019bias}. 
The GAP metric is optimized such that a model treats all demographic groups equally, i.e., its performance is (near) identical across all subgroups.
The computation details of GAP vary widely across previous work on debiasing, which impedes systematic analysis and comparison of proposed approaches.

In terms of bias mitigation, diverse debiasing methods have been proposed, including at-training-time~\citep{li-etal-2018-towards,elazar2018adversarial,shen2021contrastive}, and pre-~\citep{zhao2017men, wang2019balanced} and post-processing approaches~\citep{han2021balancing, ravfogel-etal-2020-null}. 
Although these methods have been proved effective for bias mitigation, it is challenging to reproduce results and compare methods because of inconsistencies in training strategy and model selection criterion which demonstrably affect the results.

We present \SysName, an open-source framework for bias detection and mitigation in classification tasks to address these issues.
\SysName implements a number of common debiasing approaches in a unified framework that facilitates reproducible and consistent evaluation and provides interfaces for developing new debiasing methods.
Moreover, a dataset interface supports adoption of both built-in and newly developed methods for new tasks and corpora.
For better presentation, \SysName also provides well-documented utilities for creating \LaTeX{} tables as well as performance--fairness trade-off plots for model comparison under different conditions.

\SysName is implemented based on PyTorch and is easy to use: it can be run from the command line, or imported as a package into other projects.
Model checkpoints and configuration files are saved  to keep track of hyperparameters, which is essential for reproducibility. 
To demonstrate its utility, we use \SysName to reproduce a battery of debiasing results from the recent NLP literature, and show that improved and systematic hyperparameter tuning leads to demonstrable improvements over the originally reported results. 
\SysName is released under Apache License 2.0 at  \url{https://github.com/HanXudong/fairlib}. 

\section{Fairness Criterion}
\label{sec:fairness_criterion}

\SysName includes a variety of evaluation metrics that have been proposed in previous work, as we outline in this section.

\paragraph{Group Fairness}
To evaluate whether or not a model's predictions are fair towards the protected attributes, such as gender and age, a popular way is to measure performance gap (GAP) across instances within different protected groups.
\citet{barocas-hardt-narayanan} present formal definitions of three types of group fairness criteria, 
which capture different levels of (conditional) independence between the protected attribute $g$, the target variable $y$, and the model prediction $\hat{y}$:
\begin{enumerate}
    \item \emph{independence} ($\hat{y} \perp g$), also known as \emph{Demographic Parity}~\citep{feldman2015certifying}, ensures that the positive rate of each protected group is the same;
    \item \emph{separation} ($\hat{y} \perp g | y$), also known as \emph{Equalized Odds}~\citep{hardt2016equality}, acknowledges that in many scenarios $g$ is correlated with $y$; a relaxation of this is \emph{Equal Opportunity}~\citep{hardt2016equality};
    \item \emph{sufficiency} ($y \perp g | \hat{y}$) is satisfied if the predictions are calibrated for all groups, and is also known as \emph{Test Fairness}~\citep{chouldechova2017fair}.
\end{enumerate}

\paragraph{Per-group Fairness}
Instead of measuring performance gaps across groups, per-group fairness aims to measure each subgroup's performance. 
\emph{Rawlsian Max-Min fairness}, for example, corresponds to measuring the utility (i.e., the performance) of the group with the lowest utility~\citep{rawls2001justice}. 
Similarly, \emph{Max Violation}~\citep{yang2020fairness} measures the maximum GAP across all protected groups. 

\begin{table*}[ht!]
    \centering
    \begin{adjustbox}{max width=\linewidth}
    \begin{tabular}{cll}
    \toprule
    \bf Type &  \bf Model & \bf Main Idea \\
    \midrule
    \multirow{4}{*}{\textbf{Pre-}}     
        &   \BD~\citep{zhao2017men}  &  Equalize the size of protected groups.\\
        &   \CB~\citep{wang2019balanced} &  Down-sample the majority protected group within each class.  \\
        &   \JB~\citep{NEURIPS2020_07fc15c9} &  Jointly balance the Protected attributes and classes. \\
        &   \EO~\citep{han2021balancing} & Balance protected attributes within advantage classes.\\
    \midrule
    \multirow{7}{*}{\textbf{At-}}   
        &   \Adv~\citep{li-etal-2018-towards}    &  Prevent protected attributes from being identified by the discriminator. \\
        &   \EAdv~\citep{elazar2018adversarial}   & Employ multiple discriminators for adversarial training.\\
        &   \DAdv~\citep{han2021diverse}    &  Employ multiple discriminators with orthogonality regularization for adversarial training.\\
        &   \AAdv \& \ADAdv~\citep{han2022towards} &    Enable discriminators to use target labels as inputs during training.\\
        &   \Gate~\citep{han2021balancing}   & Address protected factors with an augmented representation.\\
        &   \FairBatch~\citep{Roh2021FairBatch}  & Minimize CE loss gap though minibatch resampling. \\
        &   \FairSCL~\citep{shen2021contrastive}    & Adopt supervised contrastive learning for bias mitigation. \\
        &   \GDDiff~\citep{Shen-etal-2022-Connecting} &  Minimize the CE loss gap within each target label by adjusting the loss. \\
    \midrule
    \multirow{2}{*}{\textbf{Post-}}   
        &   \INLP~\citep{ravfogel-etal-2020-null}   &  Remove protected attributes through iterative null-space projection.\\
        &   \Gate$^{\text{soft}}$~\citep{han2021balancing}   & Adjust the prior for each group-specific component in \Gate~\citep{han2021balancing}. \\
    \bottomrule
    \end{tabular}
    \end{adjustbox}
    \caption{Built-in methods for bias mitigation, which are grouped into three types: \textbf{Pre-}processing, \textbf{At} training time, and \textbf{Post-}processing. }
    \label{tab:implemented_methods}
\end{table*}

\section{Bias Mitigation}
This section reviews the three primary types of debiasing methods, followed by a summary of bias mitigation methods implemented in \SysName.

\paragraph{Pre-processing} adjusts the training dataset to be balanced across protected groups before training, such that the input feature space is expected to be uncorrelated with the protected attributes. 
A typical line of work adopts long-tail learning approaches for debiasing, such as resampling the training set such that the number of instances within each protected group is identical~\citep{zhao2018gender, wang2019balanced,han2021balancing}.

\paragraph{At training time} introduces constraints into the optimization process for model training. 
A popular method is adversarial training, which jointly trains: (i) a discriminator to recover protected attribute values; and (ii) the main model to correctly predict the target classes while at the same time preventing protected attributes from being correctly predicted~\citep{wadsworth2018achieving, elazar2018adversarial,li-etal-2018-towards,wang2019balanced,NEURIPS2019_b4189d9d,han2021diverse}. 

\paragraph{Post-processing} aims to adjust a trained classifier according to protected attributes, such that the final predictions are fair to different protected groups. 
For example, \citet{ravfogel-etal-2020-null} iteratively projects fixed text representations from a trained model to a null-space of protected attributes. 
\citet{han2021balancing} adjust the predictions for each protected group by searching the best prior for each group-specific component.

\paragraph{Implemented Methods}
\label{sec:implemented_methods}

Table~\ref{tab:implemented_methods} lists 14 debiasing methods that are implemented in \SysName. 
It can be beneficial to employ different debiasing methods simultaneously (e.g., combine \emph{pre-processing} and \emph{training-time} methods~\cite{wang2019balanced,han2021balancing}), which
\SysName supports, and technically, every combination of these methods can be directly used without any further modifications.

\section{Model Comparison}
\label{sec:model_comparison}

In contrast to single-objective evaluation, evaluation of fairness approaches generally reports both fairness and performance simultaneously. 
Typically, no single method achieves both the best performance and fairness, making the comparison between fairness methods difficult. 
In this section, we first introduce trade-off plots for model comparison and then discuss model selection criteria that can be used for reporting numerical results. 

\begin{figure*}[ht!]
    \centering
     \begin{subfigure}[b]{0.45\textwidth}
         \centering
         \includegraphics[width=\textwidth]{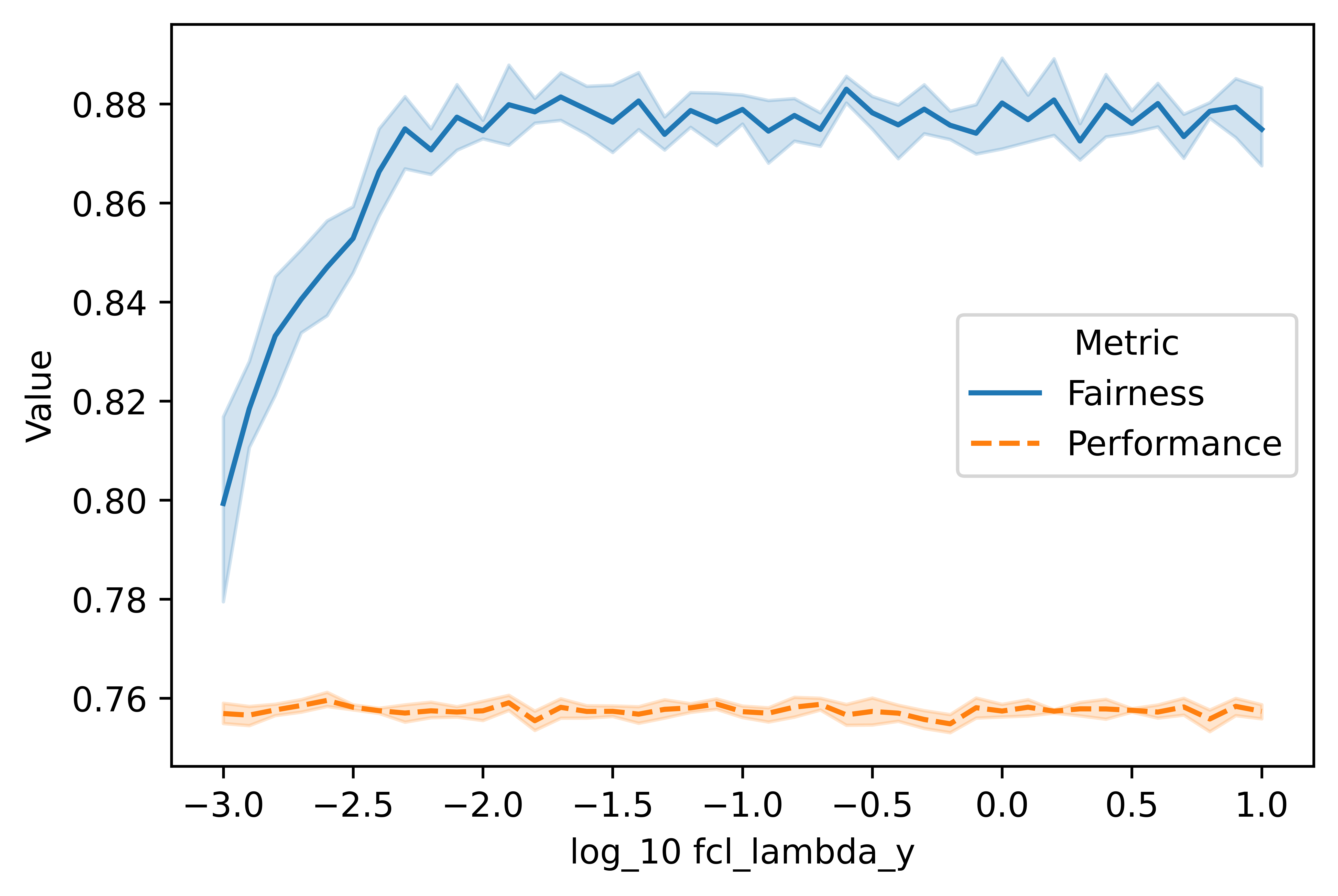}
         \caption{\Moji }
         \label{fig:moji_FSCL_hypertune}
    \end{subfigure}
    \hfill 
    \begin{subfigure}[b]{0.45\textwidth}
        \centering
        \includegraphics[width=\textwidth]{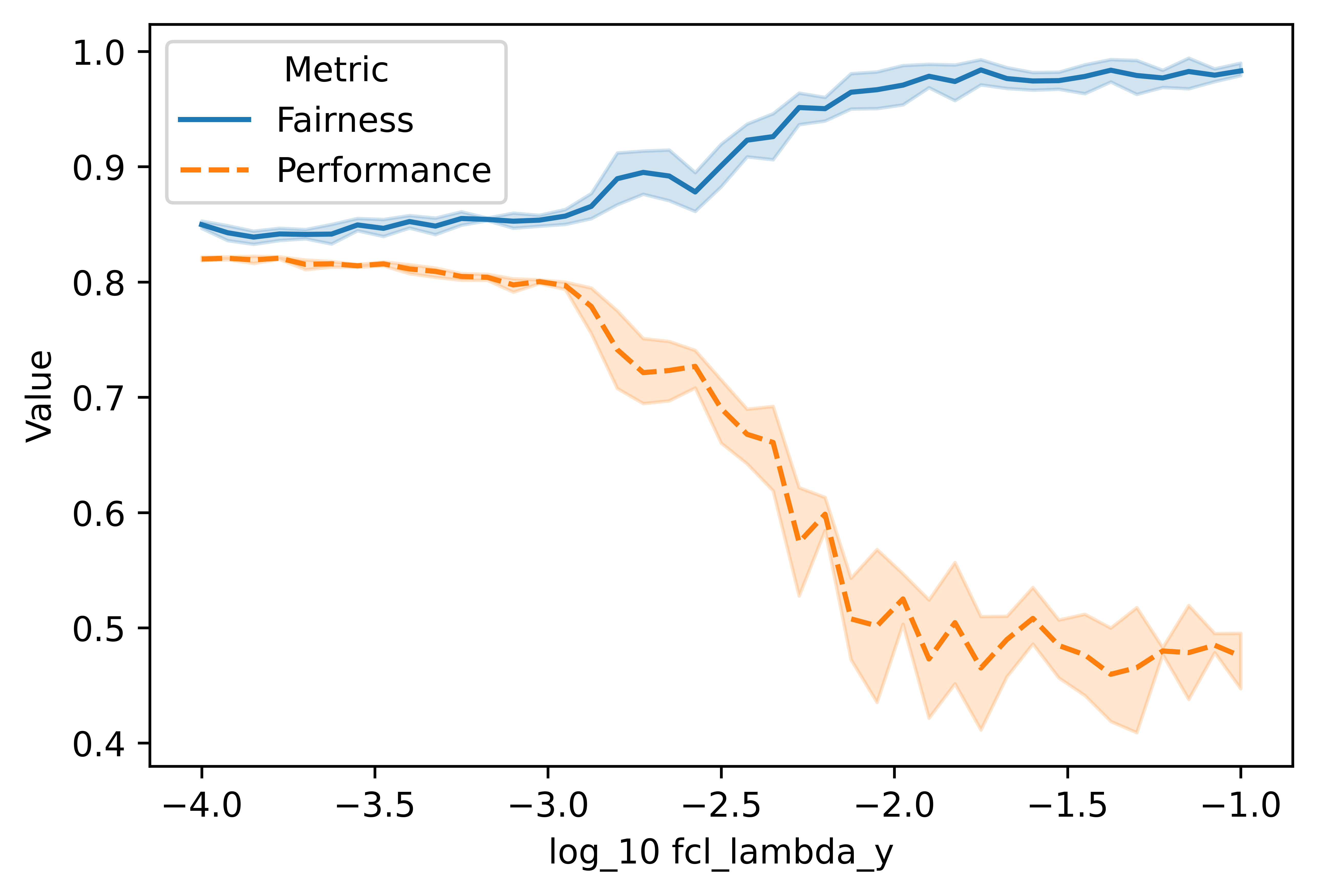}
        \caption{\Bios}
        \label{fig:bios_FSCL_hypertune}
    \end{subfigure}
    \hfill
    \caption{Tuning the same tradeoff hyperparameter of \FairSCL over two datasets. Similar trade-offs can be obtained for other debiasing methods. All figures in this paper are generated  using \SysName, including hyperparameter tuning and figure plotting. In this paper, we report the accuracy as the performance, and equal opportunity as the fairness criterion.}
    \label{fig:FSCL_hypertune}
\end{figure*}

\paragraph{Performance--fairness Trade-off} is a common way of comparing different debiasing methods without the requirement for model selection. 
Specifically, there is usually a trade-off hyperparameter for each debiasing method, which controls to what extent the final model will sacrifice performance for better fairness, such as the number of iterations for null-space projection in \INLP, or the strength of the additional contrastive losses in \FairSCL. 
Figure~\ref{fig:FSCL_hypertune} shows two trade-off plots over different values of the hyperparameter of \FairSCL applied to two different datasets (see Sec~\ref{sec:datasets} for details).
In \SysName, setting the trade-off hyperparameter to 0 for a given method degrades to the \Standard model, i.e., the naively trained model without explicit debiasing.

Typically, instead of looking at the performance--fairness with respect to different trade-off hyperparameter values, it is more meaningful to compute the maximum fairness that can be achieved by different models at a fixed performance level and vice versa.
\begin{figure*}[ht!]
    \centering
     \begin{subfigure}[b]{0.45\textwidth}
         \centering
         \includegraphics[width=\textwidth]{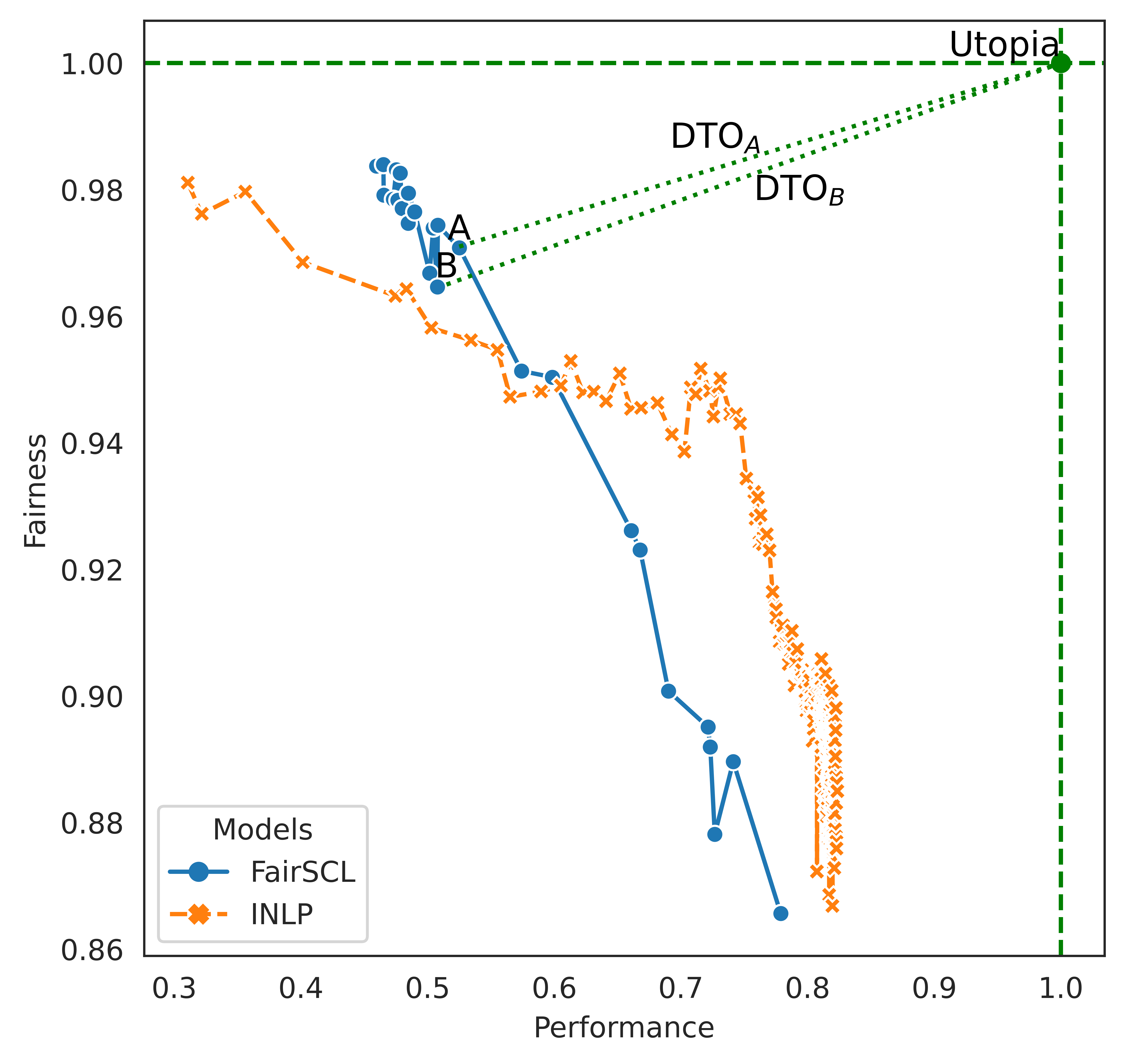}
         \caption{Trade-off
         }
         \label{fig:bios_tradeoff_FairSCL_vs_INLP}
    \end{subfigure}
    \hfill 
    \begin{subfigure}[b]{0.45\textwidth}
        \centering
        \includegraphics[width=\textwidth]{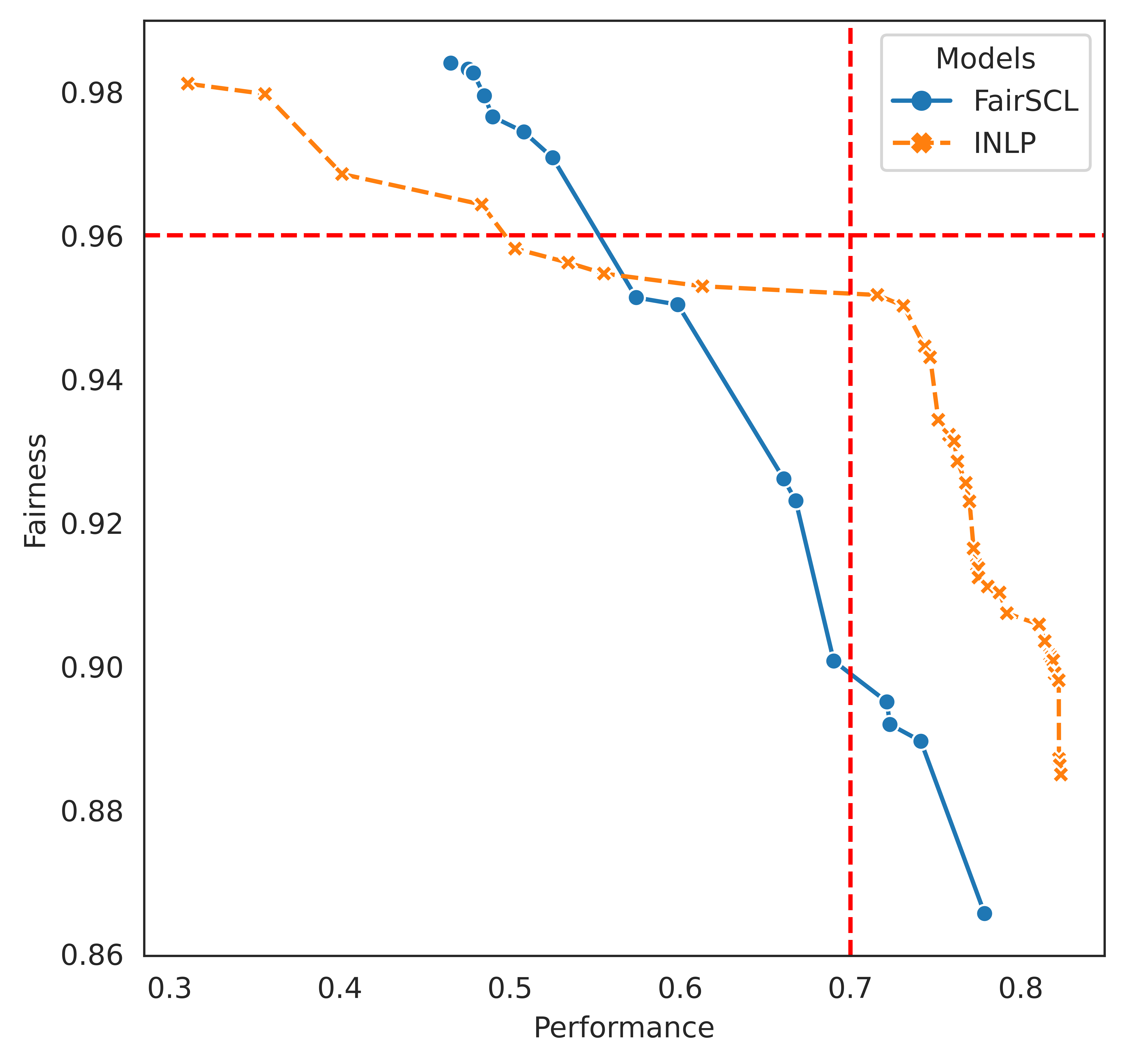}
        \caption{Pareto Trade-off}
        \label{fig:bios_tradeoff_FairSCL_vs_INLP_pareto}
    \end{subfigure}
    \hfill
    \caption{
    performance--fairness trade-offs of \FairSCL (blue points) and \INLP (orange crosses) over the \Bios dataset. The vertical and horizontal red dashed line in Figure~\ref{fig:bios_tradeoff_FairSCL_vs_INLP_pareto} are examples of constrained model selection with respect to a performance threshold of 0.7 and fairness threshold of 0.96.
    Figure~\ref{fig:bios_tradeoff_FairSCL_vs_INLP} also provides an example for \DTO. The green dashed vertical and horizontal lines denote the best performance and fairness, respectively, and their intersection point is the Utopia point. The length of green dotted lines from A and B to the Utopia point are the \DTO for candidate models A and B, respectively.
    } 
    \label{fig:bios_tradeoff_example}
\end{figure*}
Figure~\ref{fig:bios_tradeoff_FairSCL_vs_INLP} shows an example of comparing \INLP with \FairSCL over the \Bios dataset~\citep{de2019bias}, where the results are obtained by varying the hyperparameters as illustrated in Figure~\ref{fig:bios_FSCL_hypertune}.
One problem associated with such plots is that some models are strictly worse than others, i.e., achieve worse performance and fairness at the same time. 
Such models can be filtered by a simple model selection process and removed from the trade-off plots. 
Alternatively, the Pareto trade-off plot, as shown in Figure~\ref{fig:bios_tradeoff_example}, can be used. It displays the set of Pareto optimal points, where no dimension can be improved without causing a degradation in the other.

\paragraph{Model Selection} refers to the process of selecting a combination of hyperparameters that outperforms other combinations. 
In single-objective learning, model selection is based a single metric, such as the minimum loss on the dev set. 
In debiasing, however, both performance and fairness need to be considered for model selection, and a common method is \emph{Constrained Selection}, which selects the best results of one metric given a threshold of the other metric. 
This is illustrated in Figure~\ref{fig:bios_tradeoff_FairSCL_vs_INLP_pareto}, where given a performance threshold (the red vertical line), the best fairness scores that \FairSCL and \INLP can achieve are around 0.89 and 0.95, respectively. 
Similarly, the best performance scores that can be obtained given a 0.96 fairness threshold are approximately 0.53 and 0.49 for \FairSCL and \INLP, respectively.

An alternative method for model selection using Pareto frontiers is the \DTO metric~\citep{han2021balancing}, which measures the \textbf{D}istance \textbf{T}o the \textbf{O}ptimal point for candidate models  
and is widely used in multi-objective optimization~\citep{salukvadze1971concerning, marler2004survey}. 
Specifically, \DTO finds the candidate point that is closest (in terms of Euclidean distance) to the \emph{utopia point}~\cite{vincent1981optimality}. 
As shown in Figure~\ref{fig:bios_tradeoff_FairSCL_vs_INLP}, the candidate points are ordered pairs $(\text{Performance},\text{Fairness})$, and the \emph{utopia point} (optimum) represents the hypothetical system which achieves the highest-achievable performance and fairness for the dataset, shown as the green point on the top-right corner.
Lower is better for this statistic, with a minimum of 0.
Besides the ease of comparison between approaches, \DTO has the additional advantage of being minimized by the Pareto optimal points, i.e., for non-frontier candidates, there must be at least one frontier that can achieve a smaller \DTO. 

\section{\SysName Design and Architecture}
In this section, we describe the four modules of \SysName, namely data, model, evaluation, and analysis.

\subsection{Data Module}
The data module manages inputs, target labels, and protected attributes for model training and evaluation.
To enable different pre-processing debiasing methods in supporting any types of inputs, the \ex{BaseDataset} class is implemented for sampling and weight calculation based on the distribution of classes and protected attributes.
\ex{Dataset} classes inherit functionality from \ex{BaseDataset} with an additional property for loading different types of inputs.
Specifically, \SysName includes \ex{Dataset} classes for vector, matrix, and sequential inputs, to support structural inputs, image inputs, and text inputs. 
Once inputs are loaded by \ex{Dataset}, pre-processing debiasing methods are automatically applied.

Based on \ex{Dataset}, the \ex{DataLoader} class is responsible for building minibatches of samples.
Overall, steps within this module can be described as: inputs $\to$ \ex{Dataset} $\to$ \ex{DataLoader} $\to$ models, and the data module is designed for efficient data handling and pre-processing.

\subsection{Model Module}
\label{sec:model_module}

This is the core module of \SysName, which implements the \emph{At-training-time} and \emph{Post-processing} debiasing methods described in Section~\ref{sec:implemented_methods} and Table~\ref{tab:implemented_methods}.
Specifically, the methods can be applied to instances of the \ex{BaseModel} class. We provide two child classes of the \textit{BaseModel}s. First, for sequential input, we provide the implementation for BERT text classification, which combines the BERT encoder~\citep{devlin2019bert} provided by \ex{transformers} and our \ex{MLP} classifier. Second, an \ex{MLP} classifier child class of \ex{BaseModel} is implemented for structural inputs, which can be fully integrated with HuggingFace’s \ex{transformers} library.\footnote{\url{https://github.com/huggingface/transformers}}
\SysName supports combination of diverse bias mitigation methods with thousands of pre-trained models across classification tasks and data types, including text, image, and audio modalities.

\subsection{Evaluation Module}
This module implements the fairness metrics described in Section~\ref{sec:fairness_criterion} and several performance measures.
Performance measures are based on the classification evaluation metrics implemented in scikit-learn~\citep{sklearn_api}, including Accuracy, F-score, and ROC AUC.
However, no established fairness evaluation suite exists. Noting that the calculation of existing fairness metrics are all based on confusion matrices, we implemented a \ex{Evaluator} class which can calculate any confusion-matrix based fairness metrics. 
\begin{table}[t!]
    \centering
    \begin{tabular}{c|l}
    \toprule
    \bf Type &  \bf Formulation  \\
    \midrule
    \textbf{Independence} & $\frac{\tp+\fp}{\tp+\fp+\tn+\fn}$ (Positive Rate) \\
    \midrule
    \multirow{2}{*}{\textbf{Separation}}   
        &   $\frac{\tp}{\tp+\fn}$ (Recall or TPR) \\ [0.5ex]
        &   $\frac{\fp}{\fp+\tn}$ (Fall-out or FPR) \\
    \midrule
    \multirow{2}{*}{\textbf{Sufficiency}}   
        &   $\frac{\tp}{\tp+\fp}$ (Precision) \\ [0.5ex]
        &   $\frac{\tn}{\tn+\fn}$ (NPV) \\
    \bottomrule
    \end{tabular}
    \caption{Built-in fairness evaluation metrics in \SysName.}
    \label{tab:implemented_fairness_metrics}
\end{table} 
Table~\ref{tab:implemented_fairness_metrics} maps the statistical fairness criteria of Section~\ref{sec:fairness_criterion} to confusion-matrix-derived scores. The group fairness criteria (GAP) follow the difference of these scores across subgroups. 

\subsection{Analysis Module}
\label{sec:analysis_module}
This module provides utilities for model comparison as introduced in Section~\ref{sec:model_comparison}, and the two main functions are:
\begin{itemize}
    \item \ex{model\_selection\_parallel} conducts post-hoc early-stopping, which selects the best checkpoint of a model based on the desired criterion (\DTO or constrained selection). Multi-processing is supported through the \ex{joblib} library.\footnote{\url{https://joblib.readthedocs.io/en/latest/}} All results are stored for later analysis.\footnote{Experimental results are available at \url{https://github.com/HanXudong/Fair_NLP_Classification/tree/main/analysis/results}} 
    \item \ex{final\_results\_df} performs model selection consistently for different methods with the same criterion and organizes the results as a Pandas DataFrame~\citep{reback2020pandas}, which can be used to create plots and tables.
    \item \ex{interactive\_plot} takes the output DataFrame from \ex{final\_results\_df} and creates interactive plots, covering different comparison settings such as Figures~\ref{fig:bios_tradeoff_example} and~\ref{fig:trade_offs}.
\end{itemize} 
Figures~\ref{fig:FSCL_hypertune} and~\ref{fig:bios_tradeoff_example} are examples of the types of visualization supported by this module. More examples are included in Sections~\ref{sec:usage} and~\ref{sec:benchmark_experiments}.

\section{Usage}
\label{sec:usage}
In this section, we demonstrate how to use \SysName. Users can run existing models or add their own models, datasets, and metrics as needed.

\subsection{Running Existing Models}
The following command shows an example for training and evaluating a \Standard model with \SysName:
\begin{minted}
[
fontsize=\footnotesize,
breaklines
]
{bash}
python fairlib --dataset Bios_gender --emb_size 768 --num_classes 28 --encoder_architecture BERT
\end{minted}
where the task dataset, the number of distinct classes, the encoder architecture, and the dimension of embeddings extracted from the corresponding encoder need to be specified. 
The above case trains a BERT classifer over the \Bios datasets, where there are 28  professions.

In order to apply built-in debiasing methods, additional options for each debiasing methods can be added to the command-line simultaneously:
\begin{minted}
[
fontsize=\footnotesize,
breaklines
]
{bash}
python fairlib --dataset Bios_gender --emb_size 768 --num_classes 28 --encoder_architecture BERT --BT Resampling  --BTObj EO --adv_debiasing --INLP
\end{minted}
The above example employs \EO (\emph{Pre-}), \Adv (\emph{At-}), and \INLP (\emph{Post-}) at same time for a BERT classifer debiasing over the \Bios dataset.

\SysName also support \ex{YAML} configuration files with training options:
\begin{minted}
[
fontsize=\footnotesize,
breaklines
]
{bash}
python fairlib --conf_file opt.yaml
\end{minted}
which is useful for reproducing experimental results, as \SysName saves the \ex{YAML} file for each run.

\SysName can also be used be imported as a Python library, as follows:
\begin{minted}
[
baselinestretch=1.2,
fontsize=\footnotesize,
breaklines
]
{python}
from fairlib.base_options import options
from fairlib import networks

config_file = 'opt.yaml'
# Get options
state = options.get_state(conf_file=config_file)

# Init the model
model = networks.get_main_model(state)

# Training with debiasing
model.train_self()
\end{minted}
Checkpoints, evaluation results, outputs, and the configuration file are saved to the default or a specified directory.

\subsection{Performing Analysis}

As introduced in Section~\ref{sec:analysis_module}, the first step to analyze a trained model is selecting the best epoch. 
Here we provide an example for retrieving experimental results for \FairSCL, and selecting the best epoch-checkpoint:
\begin{minted}
[
baselinestretch=1.2,
fontsize=\footnotesize,
breaklines
]
{python}
from fairlib.load_results import model_selection_parallel

FairSCL_df = model_selection(
  model_id= "FSCL",
  GAP_metric_name = "TPR_GAP",
  Performance_metric_name = "accuracy",
  selection_criterion = "DTO",
  n_jobs=20,
  index_column_names = ["fcl_lambda_y", "fcl_lambda_g"],
  save_path = "FairSCL_df.pkl",)
\end{minted}
where the fairness metric is TPR GAP (corresponding to \emph{Equal Opportunity} fairness); the performance is measured with Accuracy score; the best epoch is selected based on \DTO; and the tuned trade-off hyperparameters are used as the index.
\ex{n\_jobs} is an optional argument for multi-processing, and the resulting DataFrame will be saved to the specified directory.

Assuming \ex{Bios\_gender\_results} is a Python dictionary of retrieved experimental results from the first step, indexed by the corresponding method name, we provide the following function for model comparison:
\begin{minted}
[
baselinestretch=1.2,
fontsize=\footnotesize,
breaklines
]
{python}
from fairlib.tables_and_figures import final_results_df

Bios_results = {
  "INLP":INLP_df,
  "FairSCL":FairSCL_df,}

Bios_gender_main_results = final_results_df(
  results_dict = Bios_results,
  pareto = True,
  selection_criterion = "DTO",
  return_dev = True,)
\end{minted}
where model selection is performed based on \DTO. Each method has one selected model in the resulting DataFrame, which can then be used to create tables.

If visualization is desired, users can disable the model selection by setting \ex{selection\_criterion = None}, in which case all Pareto frontiers will be included.

\subsection{Customized Datasets}

A custom dataset class must implement the \ex{load\_data} function. Take a look at this sample implementation; the split is stored in a directory \ex{self.data\_dir}. The \ex{args.data\_dir} is either loaded from the arguments \ex{--data\_dir} or from the default value. \ex{split} has three possible string values, \ex{{"train", "dev", "test"}}, indicating the split that will be loaded.

Then the \ex{load\_data} function must assign the value of \ex{self.X} as inputs, \ex{self.y} as target labels, and \ex{self.protected\_label} as information for debiasing, such as gender, age, and race.

\begin{minted}
[
baselinestretch=1.2,
fontsize=\footnotesize,
breaklines
]
{python}
from fairlib.dataloaders.utils import BaseDataset

class SampleDataset(BaseDataset):
  def load_data(self):
    # Load data from pickle file
    filename = self.split+"df.pkl"
    _Path = self.args.data_dir / filename
    data = pd.read_pickle(_Path)
        
    # Save loaded data
    self.X = data["X"]
    self.y = data["y"]
    self.protected_label = data["protected_label"] 
\end{minted}

As a child class of \ex{BaseDataset}, \emph{Pre-processing} related operations will be automatically applied to the \ex{SampleDataset}.

\subsection{Customized Models}
Recall that our current \ex{MLP} implementation (Section~\ref{sec:model_module}) can be used as a classification head for different backbone models, and the new model will support all built-in debiasing methods.

Take a look at the following example: we use BERT as the feature extractor, and then use the extracted features as the input to the MLP classifier to make predictions. 

We only need to define three functions: (1) \ex{\_\_init\_\_}, which is used to initialize the model with pretrained BERT parameters, MLP classifier, and optimizer; (2) \ex{forward}, which is the same as before, where we extract sentence representations then use the MLP to make predictions; and (3) \ex{hidden}, which is used to get hidden representations for adversarial training.

\begin{minted}
[
baselinestretch=1.2,
fontsize=\footnotesize,
breaklines
]
{python}
from transformers import BertModel
from fairlib.networks.utils import BaseModel

class BERTClassifier(BaseModel):
  model_name = 'bert-base-cased'

  def __init__(self, args):
    super(BERTClassifier, self).__init__()
    self.args = args

    # Load pretrained model parameters.
    self.bert = 
      BertModel.from_pretrained(
        self.model_name)

    # Init the classification head 
    self.classifier = MLP(args)

    # Init optimizer, criterion, etc.
    self.init_for_training()

  def forward(self, input_data, group_label = None):
    # Extract representations
    bert_output = self.bert(input_data)[1]

    # Make predictions
    return self.classifier(bert_output, group_label)
    
  def hidden(self, input_data, group_label = None):
    # Extract representations
    bert_output = self.bert(input_data)[1]
    
    return self.classifier.hidden(
      bert_output, group_label)
\end{minted}

\section{Benchmark Experiments}
\label{sec:benchmark_experiments}

To evaluate \SysName, we conduct extensive experiments to compare models implemented in \SysName with their original reported results over two benchmark datasets.

\begin{table*}[t!]
\renewrobustcmd{\bfseries}{\fontseries{b}\selectfont}
\centering
\begin{adjustbox}{max width=\linewidth}
\sisetup{
round-mode = places,
}%
\begin{tabular}{
l
S[table-format=2.2, round-precision = 2]@{\,\( \pm \)\,}S[table-format=1.2, round-precision = 2,table-number-alignment = left]
S[table-format=2.2, round-precision = 2]@{\,\( \pm \)\,}S[table-format=1.2, round-precision = 2,table-number-alignment = left]
S[table-format=2.2, round-precision = 2]
S[table-format=2.2, round-precision = 2]@{\,\( \pm \)\,}S[table-format=1.2, round-precision = 2,table-number-alignment = left]
S[table-format=2.2, round-precision = 2]@{\,\( \pm \)\,}S[table-format=1.2, round-precision = 2,table-number-alignment = left]
S[table-format=2.2, round-precision = 2]
}

\toprule
    & \multicolumn{5}{c}{\bf\Moji} & \multicolumn{5}{c}{\bf\Bios} \\
\cmidrule(lr){2-6}\cmidrule(lr){7-11}

\bf Method & \multicolumn{2}{c}{\bf Performance$\uparrow$}     &\multicolumn{2}{c}{\bf Fairness$\uparrow$} & \multicolumn{1}{c}{\bf \DTO$\downarrow$ } & \multicolumn{2}{c}{\bf Performance$\uparrow$}     &\multicolumn{2}{c}{\bf Fairness$\uparrow$} & \multicolumn{1}{c}{\bf \DTO$\downarrow$ } \\ 
\midrule
\Standard   &   72.2981 &  0.4576 &  61.1870 &    0.4356 & 47.6849 & 82.2512 & 0.2410 & 85.1071 &   0.8095 & 23.1694 \\             
\EO         &   75.3927 &  0.1433 &  87.7469 &    0.3756 & 27.4892 & 83.8326 & 0.2492 & 90.5370 &   0.9064 & \bfseries{18.73} \\
\Adv        &   75.6414 &  0.7271 &  89.3286 &    0.5623 & 26.5936 & 81.6637 & 0.2187 & 90.7356 &   0.7686 & 20.5438 \\             
\DAdv       &   75.5464 &  0.4076 &  90.4023 &    0.1218 & \bfseries{26.27} & 81.8480 & 0.1898 & 90.6376 &   0.4832 & 20.4242 \\             
\ADAdv      &   75.0163 &  0.6945 &  90.8679 &    0.1678 & 26.6004 & 81.9136 & 0.3358 & 88.9603 &   0.5943 & 21.1894 \\             
\FairBatch  &   75.0638 &  0.6012 &  90.5537 &    0.5046 & 26.6655 & 82.2382 & 0.1280 & 89.4995 &   1.2474 & 20.6335 \\
\FairSCL    &   75.7314 &  0.3441 &  87.8219 &    0.4314 & 27.1527 & 82.0594 & 0.1622 & 84.2735 &   0.8316 & 23.8577 \\
\GDDiff     &   75.2763 &  0.4999 &  89.2255 &    0.7860 & 26.9694 & 81.7773 & 0.2688 & 88.8683 &   0.9449 & 21.3537 \\
 \INLP      &   73.3433 &       0 &  85.5982 &         0 & 30.2983 & 82.3032 &        0 & 88.6249 &          0 & 21.0373 \\             

\bottomrule
\end{tabular} 
\end{adjustbox}
\caption{
Evaluation results $\pm$ standard deviation ($\%$) on the test set of sentiment analysis (\Moji) and biography classification (\Bios) tasks, averaged over 5 runs with different random seeds. 
\DTO\ is measured by the normalized Euclidean distance between each model and the ideal model, and lower is better. Due to the fact that \INLP is a \emph{post-processing} approach and its results with respect a given number of iterations are highly affected by the random seed, we only report results for 1 run. One way of getting statistics of INLP is selecting the trade-off hyperparameter of INLP for each random seed, however, this may not be a fair comparison with other methods as fixed hyperparameters have been used.}
\label{table:main_results}
\end{table*}

\begin{figure*}[ht!]
    \centering
     \begin{subfigure}[b]{0.48\textwidth}
         \centering
         \includegraphics[width=\textwidth]{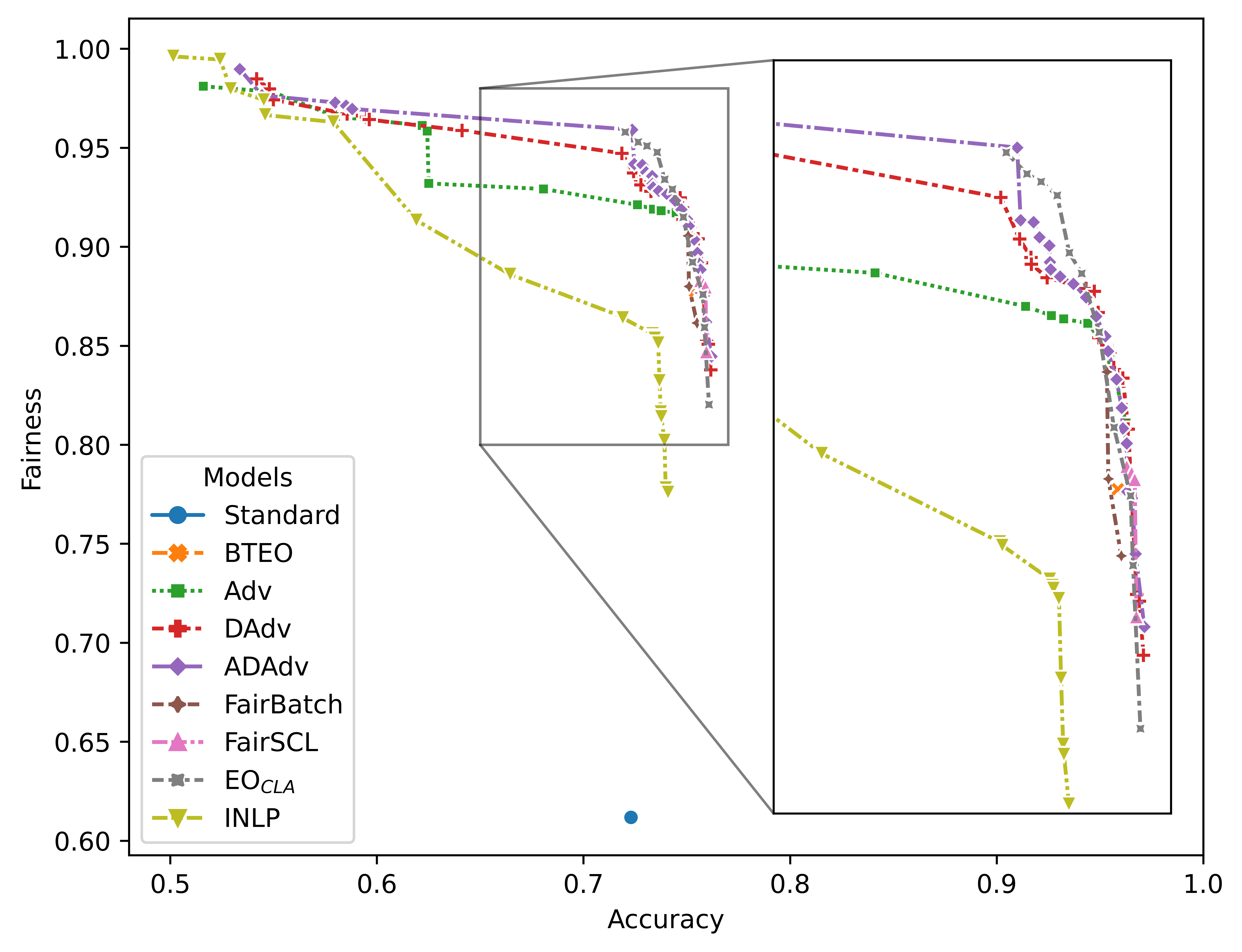}
         \caption{\Moji }
         \label{fig:Moji_trade_offs}
    \end{subfigure}
    \hfill 
    \begin{subfigure}[b]{0.48\textwidth}
        \centering
        \includegraphics[width=\textwidth]{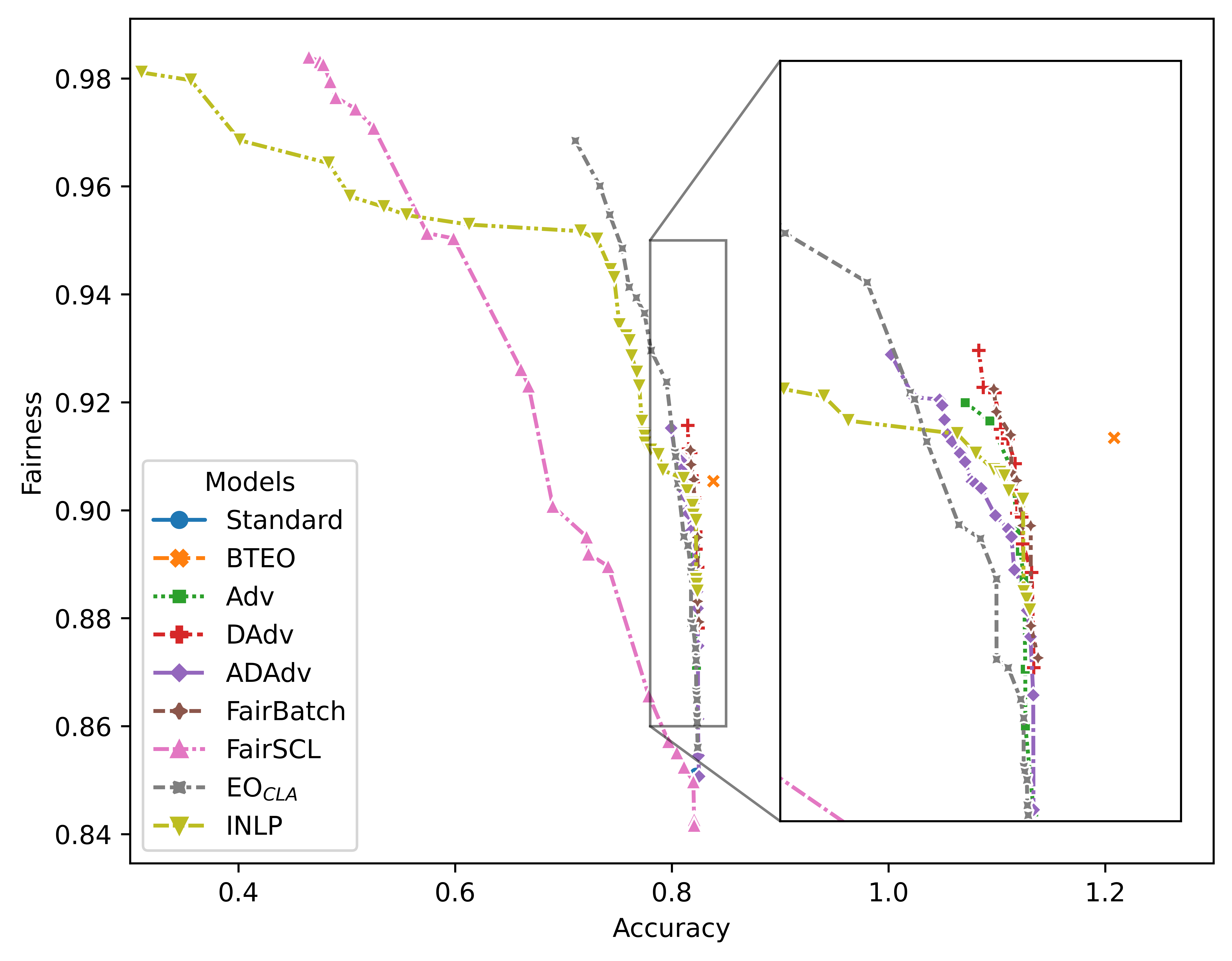}
        \caption{\Bios}
        \label{fig:Bios_trade_offs}
    \end{subfigure}
    \hfill
    \caption{Performance--fairness trade-offs of selected models over the \Moji and \Bios datasets. }
    \label{fig:trade_offs}
\end{figure*}

\subsection{Datasets}
\label{sec:datasets}
We conduct experiments over two NLP classification tasks --- sentiment analysis and biography classification --- using the same dataset splits as previous work~\citep{elazar2018adversarial,ravfogel-etal-2020-null, han2021diverse, shen2021contrastive,han2021balancing}.

\Moji:\label{sec:moji_dataset}~~
This sentiment analysis dataset was collected by \citet{blodgett-etal-2016-demographic}, and contains tweets that are either African American English (AAE)-like or Standard American English (SAE)-like.
Each tweet is annotated with a binary `race' label (based on language use: either \AAE or \SAE), and a binary sentiment score determined by (redacted) emoji contained in it.

\Bios:~~
The second task is biography classification~\citep{de2019bias}, where biographies were scraped from the web, and annotated for binary gender and 28 classes of profession.

\subsection{Evaluation Metrics}
Following~\citet{han2021balancing}, we report the overall Accuracy as the performance, and the Equal Opportunity as the fairness criterion, calculated based on the Recall gap across all protected groups.

\subsection{Models}
For illustrative purposes, we report experiments over a subset of debiasing models in Table~\ref{tab:implemented_methods}.
Specifically, the selected models are: (1)\EO~\citep{han2021balancing}, which has been shown to outperform other \emph{Pre-processing} debiasing methods; (2) \Adv~\citep{li-etal-2018-towards}; (3) \DAdv~\citep{han2021diverse}, which is the current SOTA variant of \Adv and outperforms \EAdv~\citep{elazar2018adversarial}; (4) \ADAdv~\citep{han2022towards}, which is a augmented version of \DAdv that focuses on Equal Opportunity Fairness; (5) \FairBatch~\citep{Roh2021FairBatch}; (6) \FairSCL~\citep{shen2021contrastive}; (7) \GDDiff, which is also designed for EO fairness; and (8) \INLP, which represents the \emph{Post-processing} methods.

\subsection{Experimental Results}
Table~\ref{table:main_results} summarizes the results produced by \SysName. 
Compared with previous work, \Standard, \EO, \ADAdv, \FairSCL and \GDDiff achieve similar results to the original paper. In contrast, the re-implemented \Adv, \DAdv, \FairBatch, and \INLP outperform the results reported in their original paper due to the better-designed hyperparameter tuning and model selection.\footnote{We provide further details of hyperparameter tuning in \url{https://github.com/HanXudong/fairlib/blob/main/docs/hyperparameter\_tuning.md}.}

Trade-off plots for the selected methods are shown in Figure~\ref{fig:trade_offs}.
Over the \Moji dataset (Figure~\ref{fig:Moji_trade_offs}), it can be seen that almost all methods lead to similar results, with a fairness score less than 0.9, except for \INLP, which is substantially worse than the other methods. 
As increasing the values of each model's trade-off hyperparameter (i.e., achieving better fairness at the cost of performance), \ADAdv outperforms other methods.

The trade-off plot for \Bios is quite different to \Moji: (1) \INLP becomes a reasonable choice; (2) \FairSCL does not work well over this dataset, consistent with the original paper; (3) \EO is the only method that achieves better performance than the \Standard model while increasing fairness; (4) \GDDiff could be the best choice as it achieves much better fairness than others at a comparable performance level.

\section{Related Work}

Several toolkits have been developed for learning fair AI models~\citep{bellamy2018ai, saleiro2018aequitas, bird2020fairlearn}. We discuss the two most closely-related frameworks.

The most related work to \SysName is AI Fairness 360 (AIF360), which is the first toolkit to bring together bias detection and mitigation~\citep{bellamy2018ai}.
Like \SysName, AIF360 supports a variety of fairness criteria and debiasing methods, and is designed to be extensible. The biggest difference over \SysName is that AIF360 is closely integrated with scikit-learn, and as such does not support other ML frameworks such as PyTorch.
This not only limits the applicability of AIF360 to a large portion ofa large portion of NLP and CV tasks where neural model architectures are now de rigeur, but also implies a lack of GPU support in AIF360, but also implies a lack of GPU support in AIF360.
Moreover, AIF360 only provides fundamental analysis features, such as comparing debiasing with respect to a single evaluation metric, while the analysis module of \SysName has richer features for model comparison, for example, selecting Pareto-models and interactive visualization.

The second closely-related library is FairLearn~\citep{bird2020fairlearn}, which is also targeted at assessing and improving fairness for both classification and regression tasks.
However, similar to AIF360, FairLearn is mainly developed for scikit-learn, meaning complex CV and NLP tasks are unsupported.
Additionally, FairLearn currently only supports four debiasing algorithms,\footnote{\url{https://fairlearn.org/main/user_guide/mitigation.html}} as opposed to the 14 methods supported in \SysName, providing fuller coverage of different debiasing methods.

In sum, \SysName complements exisiting tools for bias detection and improvement by: (1)~implementing a broad range of competitive debiasing approaches, with a specific focus on debiasing neural architectures which underlie many CV and NLP tasks; and (2)~comprehensive tools for interactive model comparison to help users explore the effects of different debiasing approaches.

\section{Conclusion and Future Work}

In this paper, we present \SysName, a new open-source framework for bias detection and mitigation in classification models, which implements a wide range of fairness evaluation metrics and 14 different debiasing approaches. 
With better-designed hyperparameter tuning and model selection, the reproduced models in \SysName outperform the results reported in the original work.
\SysName also has remarkable flexibility and extensibility, such that new models, debiasing methods, and datasets can be easily developed and evaluated.

For future work, we will keep \SysName up-to-date with the latest proposed debiasing approaches, and provide more model examples for other types of inputs, such as image and audio modalities.
It is worth noting that a strong assumption that has been made in this paper is that protected attributes are accessible at the training time. 
In practice, protected labels are often unavailable or only available in limited numbers. 
Sourcing protected labels can also be difficult, for reasons ranging from privacy regulations or ethical concerns, to only a small subset of users explicitly publicly disclosing protected attributes.
As such, another critical direction for future work is incorporating models that can debias with less protected labels in order to deal with real-world data, such as decoupled adversarial training~\citep{han-etal-2021-decoupling} and adversarial instance reweighting~\citep{NEURIPS2020_07fc15c9}.

\section*{Ethical Considerations}
This work provides an unified framework for measuring and improving fairness. Although \SysName assumes access to training datasets with protected attributes, this is the same data assumption made by all debiasing methods. To avoid harm and be trustworthy, we only use attributes that the user has self-identified for experiments or toy datasets. All data in this study is publicly available and used under strict ethical guidelines.

\bibliography{custom}

\begin{thebibliography}{35}
\expandafter\ifx\csname natexlab\endcsname\relax\def\natexlab#1{#1}\fi

\bibitem[{Badjatiya et~al.(2019)Badjatiya, Gupta, and
  Varma}]{badjatiya2019stereotypical}
Pinkesh Badjatiya, Manish Gupta, and Vasudeva Varma. 2019.
\newblock Stereotypical bias removal for hate speech detection task using
  knowledge-based generalizations.
\newblock In \emph{The World Wide Web Conference}, pages 49--59.

\bibitem[{Barocas et~al.(2019)Barocas, Hardt, and
  Narayanan}]{barocas-hardt-narayanan}
Solon Barocas, Moritz Hardt, and Arvind Narayanan. 2019.
\newblock \emph{Fairness and Machine Learning}.
\newblock \url{http://www.fairmlbook.org}.

\bibitem[{Bellamy et~al.(2018)Bellamy, Dey, Hind, Hoffman, Houde, Kannan,
  Lohia, Martino, Mehta, Mojsilovic et~al.}]{bellamy2018ai}
Rachel~KE Bellamy, Kuntal Dey, Michael Hind, Samuel~C Hoffman, Stephanie Houde,
  Kalapriya Kannan, Pranay Lohia, Jacquelyn Martino, Sameep Mehta, Aleksandra
  Mojsilovic, et~al. 2018.
\newblock Ai fairness 360: An extensible toolkit for detecting, understanding,
  and mitigating unwanted algorithmic bias.
\newblock \emph{arXiv preprint arXiv:1810.01943}.

\bibitem[{Bird et~al.(2020)Bird, Dud{\'\i}k, Edgar, Horn, Lutz, Milan, Sameki,
  Wallach, and Walker}]{bird2020fairlearn}
Sarah Bird, Miro Dud{\'\i}k, Richard Edgar, Brandon Horn, Roman Lutz, Vanessa
  Milan, Mehrnoosh Sameki, Hanna Wallach, and Kathleen Walker. 2020.
\newblock Fairlearn: A toolkit for assessing and improving fairness in ai.
\newblock \emph{Microsoft, Tech. Rep. MSR-TR-2020-32}.

\bibitem[{Blodgett et~al.(2016)Blodgett, Green, and
  O{'}Connor}]{blodgett-etal-2016-demographic}
Su~Lin Blodgett, Lisa Green, and Brendan O{'}Connor. 2016.
\newblock \href {https://doi.org/10.18653/v1/D16-1120} {Demographic dialectal
  variation in social media: A case study of {A}frican-{A}merican {E}nglish}.
\newblock In \emph{Proceedings of the 2016 Conference on Empirical Methods in
  Natural Language Processing}, pages 1119--1130.

\bibitem[{Buitinck et~al.(2013)Buitinck, Louppe, Blondel, Pedregosa, Mueller,
  Grisel, Niculae, Prettenhofer, Gramfort, Grobler, Layton, VanderPlas, Joly,
  Holt, and Varoquaux}]{sklearn_api}
Lars Buitinck, Gilles Louppe, Mathieu Blondel, Fabian Pedregosa, Andreas
  Mueller, Olivier Grisel, Vlad Niculae, Peter Prettenhofer, Alexandre
  Gramfort, Jaques Grobler, Robert Layton, Jake VanderPlas, Arnaud Joly, Brian
  Holt, and Ga{\"{e}}l Varoquaux. 2013.
\newblock {API} design for machine learning software: experiences from the
  scikit-learn project.
\newblock In \emph{ECML PKDD Workshop: Languages for Data Mining and Machine
  Learning}, pages 108--122.

\bibitem[{Chouldechova(2017)}]{chouldechova2017fair}
Alexandra Chouldechova. 2017.
\newblock Fair prediction with disparate impact: A study of bias in recidivism
  prediction instruments.
\newblock \emph{Big data}, 5(2):153--163.

\bibitem[{De-Arteaga et~al.(2019)De-Arteaga, Romanov, Wallach, Chayes, Borgs,
  Chouldechova, Geyik, Kenthapadi, and Kalai}]{de2019bias}
Maria De-Arteaga, Alexey Romanov, Hanna Wallach, Jennifer Chayes, Christian
  Borgs, Alexandra Chouldechova, Sahin Geyik, Krishnaram Kenthapadi, and
  Adam~Tauman Kalai. 2019.
\newblock Bias in bios: A case study of semantic representation bias in a
  high-stakes setting.
\newblock In \emph{Proceedings of the Conference on Fairness, Accountability,
  and Transparency}, pages 120--128.

\bibitem[{Devlin et~al.(2019)Devlin, Chang, Lee, and
  Toutanova}]{devlin2019bert}
Jacob Devlin, Ming-Wei Chang, Kenton Lee, and Kristina Toutanova. 2019.
\newblock Bert: Pre-training of deep bidirectional transformers for language
  understanding.
\newblock In \emph{Proceedings of the 2019 Conference of the North American
  Chapter of the Association for Computational Linguistics: Human Language
  Technologies, Volume 1 (Long and Short Papers)}, pages 4171--4186.

\bibitem[{D{\'\i}az et~al.(2018)D{\'\i}az, Johnson, Lazar, Piper, and
  Gergle}]{diaz2018addressing}
Mark D{\'\i}az, Isaac Johnson, Amanda Lazar, Anne~Marie Piper, and Darren
  Gergle. 2018.
\newblock Addressing age-related bias in sentiment analysis.
\newblock In \emph{Proceedings of the 2018 CHI Conference on Human Factors in
  Computing Systems}, pages 1--14.

\bibitem[{Elazar and Goldberg(2018)}]{elazar2018adversarial}
Yanai Elazar and Yoav Goldberg. 2018.
\newblock Adversarial removal of demographic attributes from text data.
\newblock In \emph{Proceedings of the 2018 Conference on Empirical Methods in
  Natural Language Processing}, pages 11--21.

\bibitem[{Feldman et~al.(2015)Feldman, Friedler, Moeller, Scheidegger, and
  Venkatasubramanian}]{feldman2015certifying}
Michael Feldman, Sorelle~A Friedler, John Moeller, Carlos Scheidegger, and
  Suresh Venkatasubramanian. 2015.
\newblock Certifying and removing disparate impact.
\newblock In \emph{proceedings of the 21th ACM SIGKDD international conference
  on knowledge discovery and data mining}, pages 259--268.

\bibitem[{Han et~al.(2021{\natexlab{a}})Han, Baldwin, and
  Cohn}]{han2021balancing}
Xudong Han, Timothy Baldwin, and Trevor Cohn. 2021{\natexlab{a}}.
\newblock Balancing out bias: Achieving fairness through training reweighting.
\newblock \emph{arXiv preprint arXiv:2109.08253}.

\bibitem[{Han et~al.(2021{\natexlab{b}})Han, Baldwin, and
  Cohn}]{han-etal-2021-decoupling}
Xudong Han, Timothy Baldwin, and Trevor Cohn. 2021{\natexlab{b}}.
\newblock \href {https://doi.org/10.18653/v1/2021.findings-acl.41} {Decoupling
  adversarial training for fair {NLP}}.
\newblock In \emph{Findings of the Association for Computational Linguistics:
  ACL-IJCNLP 2021}, pages 471--477.

\bibitem[{Han et~al.(2021{\natexlab{c}})Han, Baldwin, and
  Cohn}]{han2021diverse}
Xudong Han, Timothy Baldwin, and Trevor Cohn. 2021{\natexlab{c}}.
\newblock \href {https://www.aclweb.org/anthology/2021.eacl-main.239} {Diverse
  adversaries for mitigating bias in training}.
\newblock In \emph{Proceedings of the 16th Conference of the European Chapter
  of the Association for Computational Linguistics: Main Volume}, pages
  2760--2765.

\bibitem[{Han et~al.(2022)Han, Baldwin, and Cohn}]{han2022towards}
Xudong Han, Timothy Baldwin, and Trevor Cohn. 2022.
\newblock Towards equal opportunity fairness through adversarial learning.
\newblock \emph{arXiv preprint arXiv:2203.06317}.

\bibitem[{Hardt et~al.(2016)Hardt, Price, and Srebro}]{hardt2016equality}
Moritz Hardt, Eric Price, and Nati Srebro. 2016.
\newblock Equality of opportunity in supervised learning.
\newblock \emph{Advances in Neural Information Processing Systems},
  29:3315--3323.

\bibitem[{Lahoti et~al.(2020)Lahoti, Beutel, Chen, Lee, Prost, Thain, Wang, and
  Chi}]{NEURIPS2020_07fc15c9}
Preethi Lahoti, Alex Beutel, Jilin Chen, Kang Lee, Flavien Prost, Nithum Thain,
  Xuezhi Wang, and Ed~Chi. 2020.
\newblock \href
  {https://proceedings.neurips.cc/paper/2020/file/07fc15c9d169ee48573edd749d25945d-Paper.pdf}
  {Fairness without demographics through adversarially reweighted learning}.
\newblock In \emph{Advances in Neural Information Processing Systems},
  volume~33, pages 728--740.

\bibitem[{Li et~al.(2018)Li, Baldwin, and Cohn}]{li-etal-2018-towards}
Yitong Li, Timothy Baldwin, and Trevor Cohn. 2018.
\newblock \href {https://doi.org/10.18653/v1/P18-2005} {Towards robust and
  privacy-preserving text representations}.
\newblock In \emph{Proceedings of the 56th Annual Meeting of the Association
  for Computational Linguistics (Volume 2: Short Papers)}, pages 25--30.

\bibitem[{Marler and Arora(2004)}]{marler2004survey}
R~Timothy Marler and Jasbir~S Arora. 2004.
\newblock Survey of multi-objective optimization methods for engineering.
\newblock \emph{Structural and multidisciplinary optimization}, 26(6):369--395.

\bibitem[{pandas~development team(2020)}]{reback2020pandas}
The pandas~development team. 2020.
\newblock \href {https://doi.org/10.5281/zenodo.3509134} {pandas-dev/pandas:
  Pandas}.
\newblock Zenodo.

\bibitem[{Ravfogel et~al.(2020)Ravfogel, Elazar, Gonen, Twiton, and
  Goldberg}]{ravfogel-etal-2020-null}
Shauli Ravfogel, Yanai Elazar, Hila Gonen, Michael Twiton, and Yoav Goldberg.
  2020.
\newblock \href {https://doi.org/10.18653/v1/2020.acl-main.647} {Null it out:
  Guarding protected attributes by iterative nullspace projection}.
\newblock In \emph{Proceedings of the 58th Annual Meeting of the Association
  for Computational Linguistics}, pages 7237--7256.

\bibitem[{Rawls(2001)}]{rawls2001justice}
John Rawls. 2001.
\newblock \emph{Justice as fairness: A restatement}.
\newblock Harvard University Press.

\bibitem[{Roh et~al.(2021)Roh, Lee, Whang, and Suh}]{Roh2021FairBatch}
Yuji Roh, Kangwook Lee, Steven~Euijong Whang, and Changho Suh. 2021.
\newblock Fairbatch: {B}atch selection for model fairness.
\newblock In \emph{Proceedings of the 9th International Conference on Learning
  Representations}.

\bibitem[{Saleiro et~al.(2018)Saleiro, Kuester, Hinkson, London, Stevens,
  Anisfeld, Rodolfa, and Ghani}]{saleiro2018aequitas}
Pedro Saleiro, Benedict Kuester, Loren Hinkson, Jesse London, Abby Stevens, Ari
  Anisfeld, Kit~T Rodolfa, and Rayid Ghani. 2018.
\newblock Aequitas: A bias and fairness audit toolkit.
\newblock \emph{arXiv preprint arXiv:1811.05577}.

\bibitem[{Salukvadze(1971)}]{salukvadze1971concerning}
M~Ye Salukvadze. 1971.
\newblock Concerning optimization of vector functionals. i. programming of
  optimal trajectories.
\newblock \emph{Avtomat. i Telemekh}, 8:5--15.

\bibitem[{Shen et~al.(2021)Shen, Han, Cohn, Baldwin, and
  Frermann}]{shen2021contrastive}
Aili Shen, Xudong Han, Trevor Cohn, Timothy Baldwin, and Lea Frermann. 2021.
\newblock Contrastive learning for fair representations.
\newblock \emph{arXiv preprint arXiv:2109.10645}.

\bibitem[{Shen et~al.(2022)Shen, Han, Cohn, Baldwin, and
  Frermann}]{Shen-etal-2022-Connecting}
Aili Shen, Xudong Han, Trevor Cohn, Timothy Baldwin, and Lea Frermann. 2022.
\newblock Connecting loss difference with equal opportunity for fair models.
\newblock In \emph{Proceedings of the 2018 Conference of the North {A}merican
  Chapter of the Association for Computational Linguistics}, To appear.
  Association for Computational Linguistics.

\bibitem[{Vincent and Grantham(1981)}]{vincent1981optimality}
Thomas~L Vincent and Walter~Jervis Grantham. 1981.
\newblock Optimality in parametric systems(book).
\newblock \emph{New York, Wiley-Interscience, 1981. 257 p}.

\bibitem[{Wadsworth et~al.(2018)Wadsworth, Vera, and
  Piech}]{wadsworth2018achieving}
Christina Wadsworth, Francesca Vera, and Chris Piech. 2018.
\newblock Achieving fairness through adversarial learning: an application to
  recidivism prediction.
\newblock \emph{FAT/ML Workshop}.

\bibitem[{Wang et~al.(2019)Wang, Zhao, Yatskar, Chang, and
  Ordonez}]{wang2019balanced}
Tianlu Wang, Jieyu Zhao, Mark Yatskar, Kai-Wei Chang, and Vicente Ordonez.
  2019.
\newblock Balanced datasets are not enough: Estimating and mitigating gender
  bias in deep image representations.
\newblock In \emph{Proceedings of the IEEE International Conference on Computer
  Vision}, pages 5310--5319.

\bibitem[{Yang et~al.(2020)Yang, Cisse, and Koyejo}]{yang2020fairness}
Forest Yang, Moustapha Cisse, and Sanmi Koyejo. 2020.
\newblock Fairness with overlapping groups.
\newblock \emph{arXiv preprint arXiv:2006.13485}.

\bibitem[{Zhao and Gordon(2019)}]{NEURIPS2019_b4189d9d}
Han Zhao and Geoff Gordon. 2019.
\newblock \href
  {https://proceedings.neurips.cc/paper/2019/file/b4189d9de0fb2b9cce090bd1a15e3420-Paper.pdf}
  {Inherent tradeoffs in learning fair representations}.
\newblock In \emph{Advances in Neural Information Processing Systems},
  volume~32. Curran Associates, Inc.

\bibitem[{Zhao et~al.(2017)Zhao, Wang, Yatskar, Ordonez, and
  Chang}]{zhao2017men}
Jieyu Zhao, Tianlu Wang, Mark Yatskar, Vicente Ordonez, and Kai-Wei Chang.
  2017.
\newblock Men also like shopping: Reducing gender bias amplification using
  corpus-level constraints.
\newblock In \emph{Proceedings of the 2017 Conference on Empirical Methods in
  Natural Language Processing}, pages 2979--2989.

\bibitem[{Zhao et~al.(2018)Zhao, Wang, Yatskar, Ordonez, and
  Chang}]{zhao2018gender}
Jieyu Zhao, Tianlu Wang, Mark Yatskar, Vicente Ordonez, and Kai-Wei Chang.
  2018.
\newblock Gender bias in coreference resolution: Evaluation and debiasing
  methods.
\newblock In \emph{Proceedings of the 2018 Conference of the North American
  Chapter of the Association for Computational Linguistics: Human Language
  Technologies, Volume 2 (Short Papers)}, pages 15--20.

\end{thebibliography}
\bibliographystyle{acl_natbib}

\end{document}